\documentclass{article}
\usepackage{times,latexsym}
\usepackage{url}
\usepackage[T1]{fontenc}

\usepackage{amsmath,amssymb,amsthm,bbm}
\usepackage[dvipsnames]{xcolor}
\usepackage{tikz-qtree}
\usepackage{endnotes}
\usepackage{microtype}
\usepackage{graphicx}
\usepackage{subfigure}
\usepackage{booktabs} 
\usepackage{endnotes}
\usepackage{hyperref}
\usepackage{fancyhdr}
\usepackage{datetime}

\usepackage[round]{natbib}

\newcommand{\doubbr}[1]{\left[\left[#1\right]\right]}
\newcommand{\singbr}[1]{\left[#1\right]}
\newcommand{\myset}[1]{\left\{#1\right\}}
\newcommand{\paren}[1]{\left(#1\right)}

\newcommand{\expected}[2]{\underset{#1}{\E}\singbr{#2}}

\newcommand{\abs}[1]{\left|#1\right|}

\newcommand{\wh}[1]{\ensuremath{\widehat{#1}}}

\DeclareFontFamily{U}{mathx}{\hyphenchar\font45}
\DeclareFontShape{U}{mathx}{m}{n}{<-> mathx10}{}
\DeclareSymbolFont{mathx}{U}{mathx}{m}{n}
\DeclareMathAccent{\widebar}{0}{mathx}{"73}

\usepackage{pifont}

\newcommand{\si}{\ensuremath{\sigma}}

\newcommand{\al}{\ensuremath{\alpha}}
\newcommand{\be}{\ensuremath{\beta}}
\newcommand{\R}{\ensuremath{\mathbb{R}}}

\newcommand{\E}{\ensuremath{\mathbb{E}}}

\newcommand{\by}{\times}

\definecolor{light-gray}{gray}{0.80}
\definecolor{dark-gray}{gray}{0.40}

\definecolor{darkred}{rgb}{0.64, 0.0, 0.0}

\newtheorem{asm}{Assumption}[section]

\DeclareMathOperator*{\argmax}{arg\,max}

\usepackage[accepted]{icml2020}  

\newcommand{\mytitle}{
  Learning Discrete Structured Representations by
  Adversarially Maximizing Mutual Information
}

\icmltitlerunning{\mytitle}

\begin{document}

\twocolumn[
  \icmltitle{\mytitle}

\begin{icmlauthorlist}
\icmlauthor{Karl Stratos}{rutgers}
\icmlauthor{Sam Wiseman}{ttic}
\end{icmlauthorlist}
\icmlaffiliation{rutgers}{Rutgers University}
\icmlaffiliation{ttic}{Toyota Technological Institute at Chicago}
\icmlcorrespondingauthor{Karl Stratos}{karlstratos@gmail.com}
\icmlkeywords{Unsupervised Learning, Representation Learning, Mutual Information, Adversarial Training}
\vskip 0.3in
]
\printAffiliationsAndNotice{}

\begin{abstract}
We propose learning discrete structured representations from unlabeled data by maximizing the mutual information between a structured latent variable and a target variable. Calculating mutual information is intractable in this setting. Our key technical contribution is an adversarial objective that can be used to tractably estimate mutual information assuming only the feasibility of cross entropy calculation. We develop a concrete realization of this general formulation with Markov distributions over binary encodings. We report critical and unexpected findings on practical aspects of the objective such as the choice of variational priors. We apply our model on document hashing and show that it outperforms current best baselines based on discrete and vector quantized variational autoencoders. It also yields highly compressed interpretable representations.
\end{abstract}

\section{Introduction}
\label{sec:intro}

Unsupervised learning of discrete representations is appealing because
they correspond to natural symbolic representations in many domains
(e.g., phonemes in speech signals, topics in text, and objects in images).
However, working with discrete variables comes with technical challenges
such as non-differentiability and nontrivial combinatorial optimization.
Standard methods approach the problem within the framework of
variational autoencoding~\citep{kingma2014autoencoding,rezende2014stochastic} and bypass these challenges by adopting some form of
gradient approximation and possibly strong independence assumptions
\citep{bengio2013estimating,van2017neural}.

In this paper we are instead interested in a promising alternative framework
based on maximal mutual information (MMI).
Unlike autoencoding, MMI estimates a distribution over latent variables
without modeling raw signals by maximizing the mutual information
between the latent and a target variable \citep{brown1992class,bell1995information,bottleneck}.
It is well motivated as a principled approach to learning representations
that retain only predictive information and drop noise.
Its neural extensions have recently been quite successful in learning
useful continuous representations across domains \citep{Contrastive,MINE,DIM,bachman2019learning}.

We depart from these existing works on MMI in two important ways.
First, we learn discrete structured representations.
There are previous works on learning discrete representations with neural MMI \citep{IT-cotrain,PartOfSpeech},
but they only consider unstructured representations
which can transmit at most the log of the number of labels bits of information.
Breaking the log bottleneck in the discrete regime requires making encodings structured,
but it also makes exact computation intractable.
Thus the feasibility of optimizing mutual information effectively in this setting remains unclear.
We develop a tractable formulation that only requires tractable cross entropy
by a combination of mild structural assumptions and an appropriate loss function (see below).

Second, we consider a new mutual information estimator based on the difference of entropies for learning representations.
This is a crucial departure from existing works that optimize variational lower bounds \citep{poole2019variational}.
Estimators of a lower bound on mutual information have been shown to suffer fundamental limitations \citep{limit},
suggesting a need to investigate alternative estimators.
Our estimator is neither a lower bound nor an upper bound, yet it can be optimized adversarially as in
generative adversarial networks (GANs) \citep{goodfellow2014generative}.
We show for the first time that such adversarial optimization of mutual information
is a viable option for learning meaningful representations.

Our proposed discrete structured MMI is novel and largely uncharted in the literature.
An important contribution of this paper is charting practical considerations for this alien approach
by developing a concrete realization based on a structured model over binary encodings.
More specifically, the model encodes an observation into a zero-one vector of length $m$, resulting
in $2^m$ possible encodings.
We show how mutual information can be estimated efficiently by adversarial dynamic programming
with controllable Markov assumptions.
One critical and unexpected finding is that the expressiveness of a variational prior
needs to be strictly greater than that of the model (i.e., it has to be higher-order Markovian).

To demonstrate the utility of our model in a real-world problem, we apply it to unsupervised document hashing \cite{dong2019document,shen2018nash,chaidaroon2018deep}.
The task is to compress an article into a drastically smaller discrete encoding that preserves semantics.
Our model outperforms current state-of-the-art baselines based on discrete VAEs~\citep{kingma2014autoencoding,maddison2017concrete,jang2017categorical} and VQ-VAEs~\citep{van2017neural} with Bernoulli priors.
We additionally design a predictive version of document hashing in which the model is tasked with encoding a future article with the knowledge of a past article.
We find that our model achieves favorable performance with highly compressed and interpretable representations.

\section{Related Work}
\label{sec:related}
Many successful approaches to unsupervised representation learning are based on density estimation. For instance, it is now very common in natural language processing to make use of continuous representations that are learned in the process of modeling a conditional distribution $p_{Y|X}$, such as the conditional distribution of a word $Y$ given a context window $X$~\citep{mikolov2013efficient,peters2018deep,devlin2019bert}. In this case the input pair $(X,Y)$ is easily sampled from unlabeled data, by masking an observed word. There is also much work that identifies representations, often continuous, with the latent variables in an unconditional density model of $Y$~\citep{kingma2014autoencoding,rezende2014stochastic,higgins2017beta}. 

Learning representations through density estimation, however, suffers from certain limitations. First, it may be unnecessary to fully model the density of noisy, raw data when we are only interested in learning representations. Second, many standard approaches to learning discrete-valued latent representations in the context of density estimation require the use of either biased gradient estimators~\citep{bengio2013estimating,van2017neural} or high variance ones~\citep{Mnih2014,Mnih2016}.

Maximal mutual information (MMI) is a refreshingly different approach to unsupervised representation learning
in which we estimate a conditional distribution over latent representations by maximizing mutual information under these distributions.
In contrast with density estimation, there is no issue of modeling noise since the model never estimates a distribution over raw signals (i.e., there is no decoder).
The mutual information objective has been shown to produce state-of-the-art representations of images, speech, and text \citep{bachman2019learning,Contrastive}.

The focus with MMI so far has been largely limited to learning continuous representations.
Existing works on learning discrete representations with MMI
only consider unstructured one-of-$m$-labels representations \citep{IT-cotrain,PartOfSpeech} due to computational reasons.
Our main contribution is a tractable formulation for discrete structured MMI.
This involves an adversarial objective reminiscent of GANs \citep{goodfellow2014generative}
and radically different from existing MMI objectives based on variational lower bounds of mutual information \citep{poole2019variational}.
Other than tractability reasons, the choice of the objective can be theoretically motivated as avoiding
statistical limitations of estimating lower bounds on mutual information \citep{limit}.

\section{Discrete Structured MMI}
\label{sec:body}

Let $p_{XY}$ denote an unknown but samplable joint distribution over raw signals $(X, Y)$.
We assume discrete $(X, Y)$ for simplicity and relevance to our experimental setting (document hashing),
but the formulation can be easily adapted to the continuous case.
We introduce an encoder $p^\psi_{Z|Y}$ that defines a conditional distribution over
a discrete latent variable $Z$ representing the encoding of $Y$
and aim to maximize $I_\psi(X, Z)$: the mutual information between $X$ and $Z$ under $p_{XY}$ and $p^\psi_{Z|Y}$.
By the data processing inequality, the objective is a lower bound on $I(X, Y)$
and can be viewed as distilling the predictive information of $Y$ about $X$ into $Z$.

This formulation alone is meaningless since it admits the trivial solution $Z = Y$.
In order to achieve compression, there are various options.
In the information bottleneck method \citep{bottleneck}, we additionally regularize the information rate of $Z$ by simultaneously minimizing $I_\psi(Y, Z)$.
Here we advocate a more direct approach by giving an explicit budget $H_{\max}$ on the entropy of $Z$.
\begin{align*}
  \max_{\psi:\; H_\psi(Z) \leq H_{\max} }\; I_\psi(X, Z)
\end{align*}
Equivalently, we can maximize $I_\psi(X, Z)$ with a finite encoding space $\mathcal{Z}$ such that $\abs{\mathcal{Z}} \leq 2^{H_{\max}}$.
Even with small $\mathcal{Z}$ the objective is intractable because it involves marginalization over $Y$.
Using that $I_\psi(X,Z) = H_\psi(Z) - H_\psi(Z|X)$, we introduce a variational
model $q^\phi_{Z|X}$ and optimize
\begin{align}
  \max_{\psi, \phi}\; H_\psi(Z) - H_{\psi,\phi}^+(Z|X) \label{eq:mmi-lb}
\end{align}
where $H_{\psi,\phi}^+(Z|X)$ denotes the cross entropy between $p^\psi_{Z|X}$ (i.e., the distribution over $Z$ given $X$ defined under $p^\psi_{Z|Y}$ and $p_{XY}$)
and $q^\phi_{Z|X}$.
By the usual property of cross entropy, $H_{\psi,\phi}^+(Z|X)$ is an upper bound on $H_\psi(Z|X)$, hence the objective~\eqref{eq:mmi-lb} is a lower bound on $I_\psi(X,Z)$.

Unfortunately, the applicability of this variational lower bound is critically limited to settings
in which the entropy of $Z$ is tractable \citep{IT-cotrain,PartOfSpeech} or constant with respect to learnable parameters \citep{chen2016infogan,45903}.
When $\mathcal{Z}$ is small, $H_\psi(Z)$ can be easily estimated from $N$ samples $(x_1, y_1) \ldots (x_N, y_N) \sim p_{XY}$ as
\begin{align}
  \wh{H}_\psi(Z) = \sum_{z \in \mathcal{Z}} \paren{\frac{\sum_{l}  p^\psi_{Z|Y}(z|y_l)}{N}}  \log \paren{ \frac{N}{\sum_{l}  p^\psi_{Z|Y}(z|y_l)} }  \label{eq:brute}
\end{align}
But this explicit calculation is clearly infeasible for large $\mathcal{Z}$.
Furthermore, the log bottleneck on the budget $H_{\max} \leq \log \abs{\mathcal{Z}}$
then implies that it is infeasible to achieve a large information rate
using this naive formulation (e.g., even if we specify $\abs{\mathcal{Z}}$ to be a trillion we have $H_{\max} \leq 40$).

\subsection{Tractable Formulation}
\label{sec:tractable}

To allow for large $H_{\max}$, we propose to make $Z$ structured.
A simplest example of structured $Z$ is a binary vector of length $m$
which yields $\abs{\mathcal{Z}} = 2^m$ so that $H_{\max}$ can be as large as $m$.
More generally, $Z$ can be any structure whose size is exponential in some controllable integer $m$.

\subsubsection{Tractable Cross Entropy}
\label{sec:tractce}

The first key ingredient in deriving a tractable formulation is
the tractability of estimating the cross entropy between $p^\psi_{Z|X}$ and $q^\phi_{Z|X}$ from samples.

\begin{asm}
  The cross entropy $H_{\psi,\phi}^+(Z|X)$ between $p^\psi_{Z|X}$ and $q^\phi_{Z|X}$
  estimated from $N$ iid samples of $(X, Y)$
  \begin{align*}
    \wh{H}_{\psi,\phi}^+(Z|X) = -\frac{1}{N} \sum_{l=1}^N \paren{\sum_{z \in \mathcal{Z}} p^\psi_{Z|Y}(z|y_l) \log q^\phi_{Z|X}(z|x_l)}
  \end{align*}
  can be computed in time polynomial in $m$ where $\abs{\mathcal{Z}} = O(2^m)$.
  \label{asm:1}
\end{asm}

\begin{algorithm}[tb]
  \caption{CrossEntropy}
  \label{alg:xent}
  \begin{algorithmic}
    \STATE {\bfseries Input:} $p(z_i|i, z_{i-o:i-1})$ for $z_{i-o:i} \in \myset{0,1}^{o+1}$ and $i \in [m]$;
    $q(z_i|i, z_{i-o':i-1})$ for $z_{i-o':i} \in \myset{0,1}^{o'+1}$ and $i \in [m]$ where $o' \geq o$
    \STATE {\bfseries Subroutine:} $\mathrm{Forward}(p)$ that computes $\pi$ in $O(m 2^o)$ time such that $\pi(z_{i-o:i-1}|i)$ is the marginalized probability of $\bar{z} \in \myset{0,1}^{i-1}$ ending in $z_{i-o:i-1}$ under $p$ (given in the supplementary material)
    \STATE {\bfseries Output:} Cross entropy between $p$ and $q$
    \begin{align*}
      H(p,q) = - \sum_{z \in \myset{0,1}^m} p(z) \log q(z)
    \end{align*}
    \STATE {\bfseries Runtime:} $O(m 2^{o'})$
    \\\hrulefill
    \STATE {\bfseries Forward computation:} $\pi \gets \mathrm{Forward}(p)$
    \STATE {\bfseries Marginals:} For $i = 1 \ldots m$, for $z_{i-o':i} \in \myset{0,1}^{o'+1}$,
    \begin{align*}
      \mu(z_{i-o':i}|i) &\gets \pi(z_{i-o':i-o'+o-1}|i-o'+o) \\
      &\hspace{10mm} \times \paren{\prod_{j=i-o'+o}^i p(z_j|j,z_{j-o:j-1})}
    \end{align*}
    where we overwrite $\pi(z_{i-o':i-o'+o-1}|i-o'+o) = p(z_{i-o'}|i-o')$ if $o = 0$
    \STATE {\bfseries Cross entropy:} Set $H(p,q)$ as the following scalar
    \begin{align*}
      - \sum_{i=1}^m \sum_{z_{i-o':i} \in \myset{0,1}^{o'+1}} \mu(z_{i-o':i}|i) \log q(z_i|i,z_{i-o':i-1})
    \end{align*}
  \end{algorithmic}
\end{algorithm}

There is a class of structured probabilistic models with standard conditional independence assumptions such that Assumption~\ref{asm:1} holds.
For instance, in the case of $Z \in \myset{0,1}^m$, we may impose Markov assumptions and define
(with the convention $z_i = 0$ for $i < 1$)
\begin{align*}
  p^\psi_{Z|Y}(z|y) &= \prod_{i=1}^m p^\psi_{Z_i|YZ_{<i}}(z_i|y,i,z_{i-o:i-1}) \\
  q^\phi_{Z|X}(z|x) &= \prod_{i=1}^m q^\phi_{Z_i|XZ_{<i}}(z_i|x,i,z_{i-h:i-1})
\end{align*}
where $z_{i:j} = (z_i \ldots z_j)$ and $o \leq h$ are the Markov orders of $p^\psi_{Z|Y}$ and $q^\phi_{Z|X}$.
It can be easily verified that the estimate of $H_{\psi,\phi}^+(Z|X)$ based on a single sample $(x,y) \sim p_{XY}$ is
\begin{align*}
  - \sum_{i=1}^m \sum_{z_{i-h:i} } \mu(z_{i-h:i}|i,y) \log q^\phi_{Z_i|XZ_{<i}}(z_i|x,i,z_{i-h:i-1})
\end{align*}
where $\mu(z_{i-h:i}|i,y)$ is the marginal probability of the length-$(h+1)$ sequence
$z_{i-h:i} \in \myset{0,1}^{h+1}$ ending at position $i$
under the conditional distribution $p^\psi_{Z|Y}(\cdot|y)$.
These marginals can be computed by applying a variant of the forward algorithm \citep{rabiner1989tutorial} (see the supplementary material).
We give a general algorithm that computes the cross entropy between any distributions over $\mathcal{Z} = \myset{0,1}^m$
with Markov orders $o \leq o'$ in $O(m 2^{o'})$ time in Algorithm~\ref{alg:xent}.

While we focus on the choice $Z \in \myset{0,1}^m$ for concreteness, we emphasize that similar structural assumptions
can be made to consider others.
For instance, the conditional entropy of tree-structured $Z$ can be computed using a variant of the inside algorithm~\citep{hwa2000sample}.
We leave exploring other types of structure as future work.

\subsubsection{Adversarial Optimization}
\label{sec:advopt}

Assumption~\ref{asm:1} allows us to estimate the second term (cross entropy) of the objective~\eqref{eq:mmi-lb} $H_\psi(Z) - H_{\psi,\phi}^+(Z|X)$,
but it is still insufficient for estimating the objective since the first term (entropy) remains intractable.
Assumption~\ref{asm:1} only imposes conditional independence:
conditioning on the input $y$, the probability of the $i$-th value of $z$ is independent of $z_{<i-o}$ under $p^\psi_{Z|Y}$.
This independence breaks for the unconditional distribution $p^\psi_Z(z) = \E_{y \sim p_Y} [ p^\psi_{Z|Y}(z|y) ]$.
Consequently the entropy term $H_\psi(Z)$ does not decompose.
Note that while the conditional entropy $H_\psi(Z|Y)$ remains tractable, we cannot use it as a meaningful approximation of $H_\psi(Z)$
since the error is exactly $I_\psi(Y,Z)$ which is zero iff $Z$ is independent of $Y$ (i.e., vacuous encoding).

Thus we propose to introduce an additional variational model $q_Z^\theta$ to  estimate the intractable distribution $p^\psi_Z$.
We would like to make the resulting variational approximation a lower bound on $H_\psi(Z)$ so that
the objective remains maximization over all models.
Unfortunately, when entropy is large (which is our setting) meaningful lower bounds are impossible \citep{limit}.
This motivates us to again consider the cross-entropy upper bound $H_{\psi,\theta}^+(Z) \geq H_\psi(Z)$ with the following assumption.

\begin{asm}
  The cross entropy $H_{\psi,\theta}^+(Z)$ between $p^\psi_{Z}$ and $q^\theta_{Z}$
  estimated from $N$ iid samples of $Y$
  \begin{align*}
    \wh{H}_{\psi,\theta}^+(Z) = -\frac{1}{N} \sum_{l=1}^N \paren{\sum_{z \in \mathcal{Z}} p^\psi_{Z|Y}(z|y_l) \log q^\theta_Z(z)}
  \end{align*}
  can be computed in time polynomial in $m$ where $\abs{\mathcal{Z}} = O(2^m)$.
  \label{asm:2}
\end{asm}

In the binary vector setting $Z \in \myset{0,1}^m$, we can define $q^\theta_Z$ to be a Markov model of order $r \geq o$
\begin{align*}
  q^\theta_{Z}(z) &= \prod_{i=1}^m q^\theta_{Z_i|Z_{<i}}(z_i|i,z_{i-r:i-1})
\end{align*}
Then Algorithm~\ref{alg:xent} can be used to estimate $H_{\psi,\theta}^+(Z)$ in $O(m 2^r)$ time.

This gives our final objective
\begin{align}
  \max_{\psi, \phi}\; \min_{\theta}\; H_{\psi,\theta}^+(Z) - H_{\psi,\phi}^+(Z|X) \label{eq:mmi-lub}
\end{align}
which is tractable by Assumption~\ref{asm:1} and \ref{asm:2}.
Note that for any choice of $\psi$, exact optimization over $\phi$ and $\theta$ recovers the original objective $I_\psi(X,Z)$.
It can be interpreted as a simultaneously collaborative and adversarial game.
The second term (cross entropy minimization) encourages $\psi$ and $\phi$ to agree on the encoding $Z$ of $Y$.
The first term (entropy maximization) encourages $\psi$ to diversify its prediction of $Z$ but also use information from $Y$ to thwart the opponent $\theta$ who does not have access to $Y$.

A notable aspect of the objective is that it is neither an upper bound nor a lower bound on $I_\psi(X,Z)$;
we cannot guarantee that $H_{\psi,\theta}^+(Z) - H_{\psi,\phi}^+(Z|X)$ estimated from $N$ samples is larger or smaller than $I_\psi(X,Z)$.
While we lack guarantees, theoretical and empirical evidence that this bypasses limitations of lower bounds on mutual information has been shown in \citet{limit}.

\paragraph{Inference} At test time, given input $y$ we calculate
\begin{align*}
  z^* \in \argmax_{z \in \mathcal{Z}} p^\psi_{Z|Y}(z|y)
\end{align*}
and use $z^*$ as a discrete structured representation of $y$.
In the current setting in which $p^\psi_{Z|Y}$ is an order-$o$ Markov distribution over $\mathcal{Z} = \myset{0,1}^m$,
we can calculate $z^*$ in $O(m 2^o)$ time using a variant of the Viterbi algorithm \citep{viterbi1967error}.

\subsection{Practical Issues}

We give details of the proposed adversarial MMI training procedure in Algorithm~\ref{alg:adv}.
As input we assume model definitions $p^\psi_{Z|Y}$, $q^\phi_{Z|X}$, and $q^\theta_{Z}$ that satisfy Assumption~\ref{asm:1} and \ref{asm:2},
samplable $p_{XY}$, gradient update function $\mathrm{Step}$ (we use Adam for all our experiments \citep{kingma2014adam}),
and a validation task $T$.
The validation task evaluates the quality of the representation predicted by $p^\psi_{Z|Y}$
and is particularly needed since the running estimate of the adversarial objective \eqref{eq:mmi-lub} may not reflect actual progress.
We delineate certain practical issues that are important in making Algorithm~\ref{alg:adv} effective.

\begin{algorithm}[tb]
  \caption{AdversarialMMI}
  \label{alg:adv}
  \begin{algorithmic}
    \STATE {\bfseries Input:} Models $p^\psi_{Z|Y}$, $q^\phi_{Z|X}$, $q^\theta_{Z}$ satisfying Assumption~\ref{asm:1} and \ref{asm:2};
    samplable $p_{XY}$; gradient update function $\mathrm{Step}$; validation task $T$
    \STATE {\bfseries Hyperparameters:} Initialization range $\al$, batch size $N$, number of adversarial gradient steps $G$,
    adversarial learning rate $\eta'$, learning rate $\eta$, entropy weight $\be$
    \\\hrulefill
    \STATE $\psi, \phi, \theta \gets \mathrm{Unif}(-\al, \al)$
    \REPEAT
    \FOR{$S \sim p_{XY}^N$}
    \STATE \hspace{-3mm} For $G$ times: $\theta \gets \mathrm{Step}\paren{\wh{H}_{\psi,\theta}^+(Z), \theta, \eta'}$
    \STATE \hspace{-3mm} $\psi, \phi \gets \mathrm{Step}\paren{\wh{H}_{\psi,\phi}^+(Z|X) - \be \wh{H}_{\psi,\theta}^+(Z), \myset{\psi, \phi}, \eta}$
    \ENDFOR
    \UNTIL{$T(\psi, \phi, \theta)$ stops improving}
  \end{algorithmic}
\end{algorithm}

\paragraph{Expressive variational prior}
We find that it is critical to make the variational prior $q^\theta_{Z}$ strictly more expressive than the posterior $p^\psi_{Z|Y}$.
That is, there exist distributions over $\mathcal{Z}$ that can be modeled by $q^\theta_{Z}$ but not by $p^\psi_{Z|Y}$ (conditioning on any $y$).
In the context of Markov models over binary vectors $Z \in \myset{0,1}^m$,
this means the Markov order $r$ of $q^\theta_Z$ is strictly greater than the Markov order $o$ of $p^\psi_{Z|Y}$
(Algorithm~\ref{alg:xent} allows for any $r \geq o$).
Recall that $Z_1 \ldots Z_m$ are conditionally independent under $p^\psi_{Z|Y}$ but not independent under $p^\psi_Z$ (Section~\ref{sec:advopt}).
Thus we must model $p^\psi_Z$ using a distribution $q^\theta_Z$ that is strictly more powerful than $p^\psi_{Z|Y}$.
The benefit of this explicit joint entropy maximization
is suggested in experiments later in which we show that our approach is more effective at learning representations
than discrete VAEs or VQ-VAEs (which do not explicitly maximize entropy) as $m$ becomes larger.

We also find it helpful to overparameterize $q^\theta_Z$.
We use a feedforward network with $\gg m 2^r$ parameters and a tunable number of ReLU layers to define the distribution.

\paragraph{Aggressive inner-loop optimization}
The adversarial objective~\eqref{eq:mmi-lub} reduces to the non-adversarial objective~\eqref{eq:mmi-lb}
if the inner minimization over $\theta$ is solved exactly.
We find it important to mimic this by taking multiple ($G$) gradient steps for $\theta$
with large learning rate $\eta'$ before taking a gradient step for $\myset{\psi, \phi}$.
Aggressive inner-loop optimization has been shown helpful in other contexts such as VAEs \citep{he19iclr}.
Note that $\theta$ is still carried across batches and not learned from scratch at every batch.

\paragraph{Entropy weight}
Finally, we find it useful to introduce a tunable weight $\be \geq 1$ for the entropy term, akin to the weight for KL divergence in $\be$-VAEs \citep{higgins2017beta}.
Note that optimizing this weighted objective is equivalent to optimizing $I_\psi(X, Z) + (\be - 1) H_\psi(Z)$.
The weight can be used to determine a task-specific trade-off between predictiveness and diversity in $Z$.


\section{Experiments}
\label{sec:experiments}

We now study empirical aspects of our proposed adversarial MMI approach (henceforth AMMI) with extensive experiments.\footnote{Code: \url{https://github.com/karlstratos/ammi}}
We consider unsupervised document hashing \citep{chaidaroon2018deep} as a main testbed for evaluating the quality of learned representations.
The task is to compress an article into a drastically smaller discrete encoding that preserves semantics
and formulated as an autoencoding problem.
To study methods in a predictive setting, we also develop a variant of this task in which the representation of an article is learned to be predictive of the encoding of a related article.

\subsection{Unsupervised Document Hashing}

Let $Y$ be a random variable corresponding to a document.
The goal is to learn a document encoder $q_{Z|Y}$ that defines a conditional
distribution over binary hashes $Z \in \myset{0, 1}^m$.
The quality of document encodings is evaluated by the average top-100 precision.
Specifically, given a document at test time,
we retrieve 100 nearest neighbors from the training set under the encoding
measured by the Hamming distance
and check how many of the neighbors have overlapping topic labels (thus we assume annotation only for evaluation).

In the literature this is typically approached as an autoencoding problem in which $q_{Z|Y}$ is estimated by
maximizing the evidence lower bound (ELBO) on the marginalized log likelihood of training documents
\begin{align}
  \max_{q_{Z|Y}, p_{Y|Z}}\; \expected{\substack{y \sim p_Y \\ z \sim q_{Z|Y}}}{\log p_{Y|Z}(y|z)} - D_{\mathrm{KL}}\paren{q_{Z|Y} \big|\big| p_Z} \label{eq:elbo}
\end{align}
where $p_Z$ is a fixed prior suitable for the task.
For example, the current state-of-the-art model (BMSH) defines $p_Z$ as a mixture of Bernoulli distributions \cite{dong2019document}.
Here $q_{Z|Y}$ is treated as a variational model that estimates the intractable posterior
$p_{Z|Y}$ under the model $p_{YZ}(y, z) = p_{Y|Z}(y|z) p_Z(z)$.

In contrast, we propose to learn a document encoder $p^\psi_{Z|Y}$ by
the following adversarial formulation of the mutual information between $Y$ and $Z$:
\begin{align}
  \max_{\psi}\; \min_{\theta}\; H_{\psi,\theta}^+(Z) - H_{\psi}(Z|Y) \label{eq:mmi-lub-single}
\end{align}
where $q^\theta_Z$ is a variational model that estimates the intractable prior $p^\psi_Z$ under the model $p^\psi_{YZ}(y, z) = p^\psi_{Z|Y}(z|y) p_Y(y)$.
This can be seen as a single-variable variant of the more general objective~\eqref{eq:mmi-lub} in which $X=Y$ and $\phi$ is tied with $\psi$.
Note the absence of a decoder (Section~\ref{sec:related}).

\subsubsection{Models}
\label{sec:models}

\paragraph{BMSH}
We follow the standard setting in BMSH for all our models \citep{dong2019document}.
The raw document representation $y$ is a high-dimensional TFIDF vector computed from preprocessed corpora (TMC, NG20, and Reuters) provided by \citet{chaidaroon2018deep}.
We aim to learn an $m$-dimensional binary vector representation $z \in \myset{0, 1}^m$ where we vary the number of bits $m=16, 32, 64, 128$.
All VAE-based models compute a continuous embedding $\tilde{z}$ by feeding $y$ through a feedforward layer (FF)
and apply some discretization operation on $\tilde{z}$ to obtain $z$.
BMSH computes $\tilde{z} = \si(\mathrm{FF}(y)) \in \R^m$ and samples $z \sim \mathrm{Bernoulli}(\tilde{z})$
from which $y$ is reconstructed.\footnote{We refer to \citet{dong2019document} for details of the decoder and the Bernoulli mixture prior since they are not needed for AMMI.}
The ELBO objective~\eqref{eq:elbo} is optimized by straight-through estimation \citep{bengio2013estimating}.\footnote{That is, the decoder receives  $(z - \tilde{z}).\mathrm{detach}() + \tilde{z}$ to backpropagate gradients directly to $\tilde{z}$.}

\paragraph{DVQ}
We are interested in comparing AMMI to mainstream discrete representation learning techniques.
Thus we additionally consider vector quantized VAEs (VQ-VAEs) which
substitute sampling with a nearest neighbor lookup and are shown to be useful in many tasks \citep{van2017neural,razavi2019generating}.
In particular, we adopt the decomposed vector quantization (DVQ) proposed in \citet{kaiser2018fast}.
The model learns $m$ codebooks $C_1 \ldots C_m \in \R^{2 \by D}$ where the row index in $C_i$ corresponds to $z_i \in \myset{0, 1}$.
The encoder computes $\tilde{z} = \mathrm{FF}(y) \in \R^{mD}$
and quantizes the $i$-th segments in $\tilde{z}$ against $C_i$ (implicitly yielding $z$)
from which the decoder reconstructs $y$.
DVQ is trained by minimizing the reconstruction loss and the vector quantization loss.
It can be seen as optimizing the ELBO objective~\eqref{eq:elbo} with a uniform prior over $Z$ and a point-mass posterior $q_{Z|Y}$.

\paragraph{AMMI}
Our model consists of an encoder $p^\psi_{Z|Y}$ and a variational prior $q^\theta_Z$,
which are respectively parameterized by order-$o$ and order-$r$ Markov models over $\myset{0,1}^m$.
The encoder computes $p^\psi_{Z|Y}(\cdot|y) = \si(\mathrm{FF}(y)) \in \R^{m2^o}$ which gives
the model's probability of $z_i=1$ conditioning on each value of $z_{i-o:i-1}$ for every $i \in [m]$.
Similarly, the prior computes $q^\theta_Z = \si(\mathrm{FF}(\Theta)) \in \R^{m2^r}$
where $\Theta \in \R^{m \by H}$ is a learnable embedding dictionary with dimension $H$.
We can then use Algorithm~\ref{alg:xent} to calculate
\begin{align*}
  \wh{H}_{\psi,\theta}^+(Z) &= \mathrm{CrossEntropy}\paren{p^\psi_{Z|Y}(\cdot|y), q^\theta_Z} \\
  \wh{H}_{\psi}(Z|Y) &= \mathrm{CrossEntropy}\paren{p^\psi_{Z|Y}(\cdot|y), p^\psi_{Z|Y}(\cdot|y)}
\end{align*}
where in practice we use a batch of samples $y_1 \ldots y_N$ to estimate these quantities.
The algorithm can be batched efficiently.

\subsubsection{Hyperparameter Tuning}

All hyperparameters for AMMI are shown in Algorithm~\ref{alg:adv}.
We perform random grid search on the validation portion of each dataset.
We find that an effective range of values is: initialization $\al = 0.1$,
batch size $N \in \myset{16, 32, 64, 128}$,
adversarial step $G \in \myset{1, 2, 4}$,
adversarial learning rate $\eta' \in \myset{0.03, 0.01, 0.003, 0.001}$,
learning rate $\eta \in \myset{0.03, 0.01, 0.003, 0.001, 0.0003, 0.0001}$,
and entropy weight $\be \in \myset{1, 1.5, 2, 2.5, 3, 3.5}$.
We similarly perform random grid search on all hyperparameters of BMSH and DVQ.
We use an NVIDIA Quadro RTX 6000 with 24GB memory.

The Markov orders $o$ and $r$ of the encoder and the prior are also controllable hyperparameters in AMMI.
We find that setting $o = 0$ is sufficient for this task (i.e., bits are independent conditioning on the document).
But the choice of $r$ is crucial, as we show below.

\begin{figure}[t!]
\vskip 0.2in
\begin{center}
  \centerline{
    \includegraphics[width=\columnwidth/2]{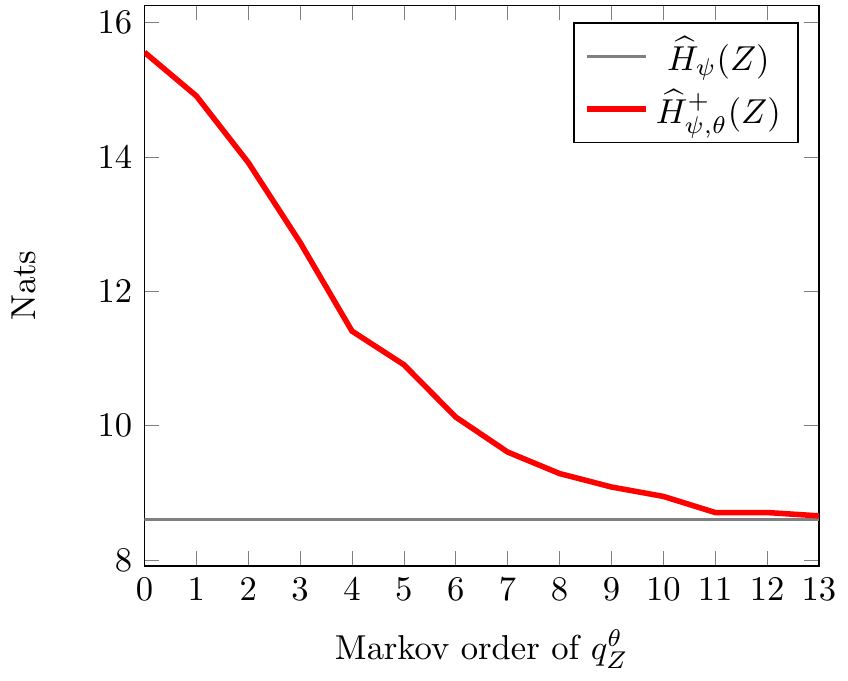}
    \includegraphics[width=\columnwidth/2]{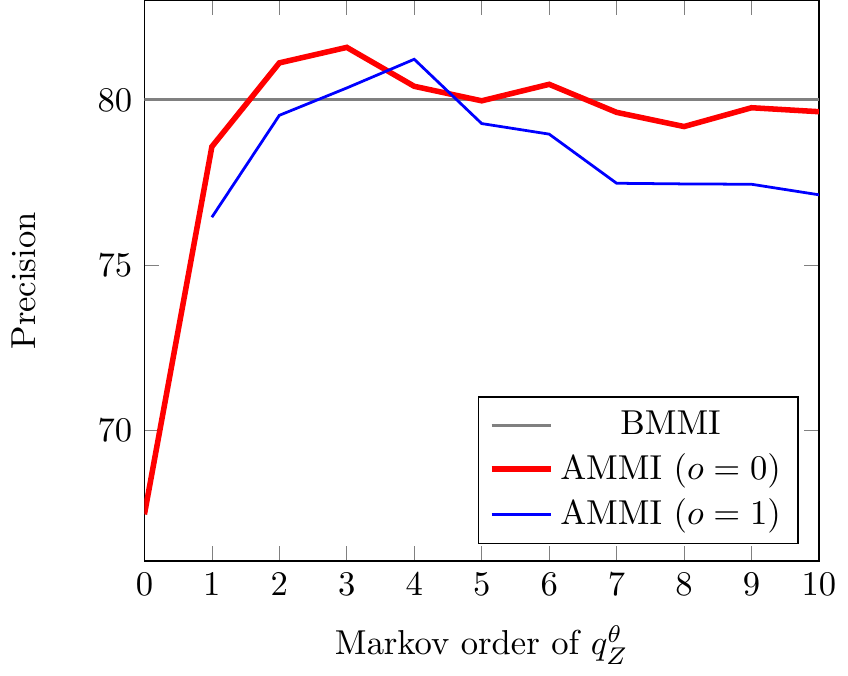}
  }
  \hspace{8mm}(a) \hspace{35mm} (b)
  \caption{Behavior of the model using different Markov orders $r$ for the variational prior $q^\theta_Z$;
    we use $m=16$ bits and $o=0$ as the Markov order of $p^\psi_{Z|Y}$.
    For each choice of $r$ all hyperparameters of the variational model are fully optimized.
    (a) shows the cross-entropy upper bound estimated on a fixed batch of samples.
    (b) shows the validation precision on Reuters.
    The blue line additionally shows the result using $o=1$.
    The gray line corresponds to the setting in which we calculate the entropy by brute-force.}
\label{fig:order}
\end{center}
\vskip -0.2in
\end{figure}

\begin{table*}[t]
  \caption{Results on unsupervised document hashing. For each dataset we show test precisions of the top 100 retrieved documents using $\myset{16, 32, 64, 128}$-bit binary vector encoding.
    See the main text for task and model descriptions.}
  \label{tab:main}
  \vskip 0.15in
  \begin{center}
    \begin{footnotesize}
      \begin{sc}
        \begin{tabular}{l|cccc|cccc|cccc|c}
          \toprule
          Data   & \multicolumn{4}{|c|}{TMC}        & \multicolumn{4}{|c|}{NG20}       & \multicolumn{4}{|c|}{Reuters}     & Avg \\
                 & 16b & 32b & 64b & 128b & 16b & 32b & 64b & 128b & 16b & 32b & 64b & 128b  &  \\
          \midrule
          BOW    & \multicolumn{4}{|c|}{50.86}      & \multicolumn{4}{|c|}{9.22}       & \multicolumn{4}{|c|}{57.62}       & 39.23 \\
          \midrule
          LSH    & 43.93 & 45.14 & 45.53 & 47.73  & 5.97  & 6.66  & 7.70  & 9.49   & 32.15 & 38.62 & 46.67 & 51.94   & 31.79 \\
          S-RBM	 & 51.08 & 51.66 & 51.90 & 51.37  & 6.04  & 5.33  & 6.23  & 6.42   & 57.40 & 61.54 & 61.77 & 64.52   & 39.61 \\
          SpH    & 60.55 & 62.81 & 61.43 & 58.91  &32.00  & 37.09 & 31.96 & 27.16  & 63.40 & 65.13 & 62.90 & 60.45   & 51.98 \\
          STH	& 39.47 &41.05 & 41.81 &41.23& 52.37 &58.60 & 58.06 & 54.33 & 73.51 & 75.54 & 73.50 & 69.86 & 56.61 \\
          VDSH & 68.53 & 71.08 & 44.10 & 58.47 & 39.04 &43.27 & 17.31 & 5.22 & 71.65 & 77.53 & 74.56 & 73.18 & 53.66 \\
          NASH & 65.73 & 69.21 & 65.48 & 59.98 & 51.08 & 56.71 & 50.71 & 46.64 & 76.24 & 79.93 & 78.12 & 75.59 & 64.62 \\
          GMSH & 67.36 & 70.24 & 70.86 & 72.37 & 48.55 & 53.81 & 58.69 & 55.83 & 76.72 & 81.83 & 82.12 & 78.46 & 68.07 \\
          DVQ & \textbf{71.47} & 73.27  &  75.17 & 76.24 & 47.23 & 54.45  & 58.77 & 62.10 & 79.57 & 83.43 & 83.73 & \textbf{86.27} & 70.98 \\
          BMSH & 70.62 & \textbf{74.81} & 75.19 & 74.50 & \textbf{58.12} & \textbf{61.00} & 60.08 & 58.02 & 79.54 & 82.86 & 82.26 & 79.41 & 71.37 \\
          \midrule
          AMMI  & 70.96 & 74.16 & \textbf{75.22} & \textbf{76.27} & 55.18 & 59.56 & \textbf{63.98} & \textbf{66.18} & \textbf{81.73} & \textbf{84.46} & \textbf{85.06} & 86.02 & \textbf{73.23} \\  
          BMMI & 70.52 &      &       &       & 49.74 &  & & & 79.97 & & & &  \\
          \bottomrule
        \end{tabular}
      \end{sc}
    \end{footnotesize}
  \end{center}
  \vskip -0.1in
\end{table*}

\subsubsection{Importance of the Markov Order of the Variational Prior}

We first examine the feasibility of variational approximation of the prior.
To this end, we consider the small-bit setting $m=16$ in which we can
explicitly enumerate $2^{16}$ values of $Z$ to estimate the entropy $H_\psi(Z)$ using equation~\eqref{eq:brute}.
In this case the objective becomes non-adversarial.
We refer to this model as BMMI (brute-force MMI) which only consists of $p^\psi_{Z|Y}$
trained by $\max_{\psi} H_\psi(Z) - H_{\psi}(Z|Y)$.

Figure~\ref{fig:order} shows two experiments that illustrate the importance of the Markov order $r$ of the variational prior $q^\theta_Z$.
First, we fix a BMMI with order $o=0$ partially trained on Reuters and a random batch of samples
to calculate the empirical entropy $\wh{H}_\psi(Z)$ by brute-force.
Then for each choice of $r = 0, 1, \ldots, 13$,
we minimize the empirical cross entropy $\wh{H}_{\psi,\theta}^+(Z)$ between $p^\psi_{Z|Y}$ and $q^\theta_Z$ over $\theta$ with full hyperparameter tuning.
Figure~\ref{fig:order}(a) shows that we need $r \gg o$ to achieve a realistic estimate of the empirical entropy.
The necessary value of $r$ clearly depends on $p^\psi_{Z|Y}$: $r > 10$ yields an accurate estimate for the partially trained BMMI used in this experiment.

Next, we examine the best achievable validation precision on Reuters across different values of $r$.
We perform full hyperparameter tuning for BMMI ($o=0$) and for AMMI with $r = 0, 1, \ldots, 10$.
We see that the performance is poor for $r=o=0$, but it quickly becomes competitive as $r > 0$ and even surpasses the performance of BMMI.
We hypothesize that the adversarial formulation has beneficial regularization effects in addition to making the objective tractable for large $m$:
we leave a deeper investigation of this phenomenon as future work.
We use $r = 3$ for all our experiments.

\subsubsection{Results}

Table~\ref{tab:main} shows top-100 precisions on the test portion of each dataset TMC, NG20, and Reuters using $m=16, 32, 64, 128$ bits.
Baselines include locality sensitive hashing (LSH) \citep{datar2004locality}, stack restricted Boltzmann machines (S-RBMs) \citep{hinton2012practical},
spectral hashing (SpH) \citep{weiss2009spectral}, self-taught hashing (STH) \citep{zhang2010self},
variational deep semantic hashing (VDSH) \citep{chaidaroon2018deep}, and neural architecture for semantic hashing (NASH) \citep{shen2018nash},
as well as BMSH \citep{dong2019document} and DVQ described in Section~\ref{sec:models}.
The naive baseline BOW refers to the bag-of-words representation $\mathrm{BOW}(y) \in \myset{0,1}^{\abs{V}}$
that indicates presence of words in document $y$.\footnote{The vocabulary size $\abs{V}$ is $20000$, $9988$, and $7164$ for TMC, NG20, and Reuters.}

We see that AMMI performs favorably to current state-of-the-art methods, yielding the best average precision across datasets and settings.
In particular, the precision of AMMI is significantly higher than the best previous result given by BMSH when $m$ is large.
With 128 bits, AMMI achieves 76.27 vs 74.50, 66.18 vs 58.02, and 86.02 vs 79.41 on TMC, NG20, and Reuters.
We hypothesize that this is partly due to the explicit entropy maximization in AMMI that considers all bits jointly via dynamic programming.
While this is implicitly enforced in the mixture prior in BMSH, direct entropy maximization seems to be more effective.

In the case of 16 bits, we also report precisions of BMMI that estimates entropy exactly by brute-force (it is computationally intractable to train BMMI with larger than $16$ bits).
We see that AMMI is again able to achieve better results potentially due to regularization effects.
Finally, we observe that the newly proposed DVQ baseline is quite competitive with BMSH and also achieves higher precision when $m$ is large;
we suspect that the decomposed encoding allows the model to make better use of multiple codebooks as reported in \citet{kaiser2018fast}.

\begin{table*}[t!]
  \caption{Qualitative analysis of AMMI document representations learned by predictive document hashing.
    For each considered document (top row), we show documents with Hamming distance at least 1, 5, 10, 20, 50, and 90 under the representations
    to examine their semantic differences.}
  \label{tab:drift}
  \vskip 0.15in
  \begin{center}
    \begin{tiny}
      \begin{tabular}{l|l}
        \toprule
        Distance   & Document \\
        \midrule
        0 & \textbf{O.J. Simpson lashed out at the family of the late Ronald Goldman, a day after they won the rights to Simpson's canceled "If I Did It" book about the slayings of Goldman } \\
        1 & News Corp. on Monday announced that it will cancel the release of a new book by former American football star O.J. Simpson and a related exclusive television interview \\
        5 & Phil Spector's lawyers have asked the judge to tell jurors they must find the record producer either guilty or not guilty of murder with no option to find lesser offenses \\
        10 & Sen. Ted Stevens' defense lawyer bore in on the prosecution's chief witness on Tuesday, portraying him to a jury as someone who betrayed a longtime friend to protect his fortune. \\
        20 & Words that cannot be said on American television are not often uttered at the U.S. Supreme Court, at least not by high-priced lawyers and the justices themselves.  \\
        50 & Cols 1-6: Sending a strong message that the faltering economy will be his top focus, President-elect Barack Obama on Friday urged Congress to pass an economic stimulus package \\
        90 & President Hu Jintao's upcoming visits to Latin America and Greece would boost bilateral relations and deepen cooperation \\
        \midrule
        0 &  \textbf{Ukrainian President Leonid Kuchma had a meeting on Monday evening with Polish President Alexander Kwasniewski and Lithuanian President Valdas Adamkus} \\
        1 & Radical Ukrainian opposition figure Yulia Timoshenko Wednesday ventured into the hostile eastern mining bastion of Prime Minister Viktor Yanukovich \\
        5 & Ukrainian President Viktor Yushchenko was forced into an emergency landing Thursday and seized the aircraft of his bitter political foe, Prime Minister Yulia Tymoshenko \\
        10 & On a clearing in this disputed city, where enemy homes were bulldozed after the conflict in August, Mayor Yuri M. Luzhkov promised this month to build a new neighborhood \\
        20 & Barack Obama is the "American Gorbachev" who will ultimately destroy the United States, militant Russian nationalist Vladimir Zhirinovsky said Tuesday. \\
        50 & Ministers from Pacific Rim nations warned Thursday that imposing trade barriers in reaction to the global economic downturn would only deepen the crisis. \\
        90 & We shall move the following graphics: US IRAQ QAEDA Graphic with portraits of Osama bin Laden and Colin Powell, examining US claims that the latest bin Laden tape reinforces \\
        \midrule
        0 & \textbf{NASCAR has a new rivalry: Carl Edwards vs. Kyle Busch. Edwards called the latest installment payback and Busch promised that retribution will come down the road.} \\
        1 & Penske Racing teammates Ryan Briscoe and Helio Castroneves filled the front-row Friday for Edmonton Indy, repeating their 1-2 finish in last week's IndyCar race \\
        5 & Brazilian race-car driver Helio Castroneves upset Spice Girl Melanie Brown to capture the fifth "Dancing With the Stars" mirrorball trophy in the television dance competition. \\
        10 & Audi overcame the challenge of two Peugeot cars and wet racing conditions Sunday to win the 24 Hours of Le Mans for the fourth straight year.  \\
        20 & The president of cycling's governing body Monday insists the doping problems in his sport do not threaten its place in the Olympics.  \\
        50 & Robert Pattinson, who stars as the vampire heartthrob Edward Cullen in the forthcoming movie "Twilight," stepped onto a riser at the King of Prussia Mall\\
        90 & Australian Prime Minister John Howard reshuffled his cabinet Tuesday, appointing Education Minister Brendan Nelson to the defence portfolio. \\
        \midrule
        0 & \textbf{How did that Van Halen song go? ``I found the simple life ain't so simple ... '' The iconic Los Angeles metal band is set to be inducted Monday night into } \\
        8 & The boys from Van Halen, most notably mercurial guitarist Eddie Van Halen, showed up as promised at a news conference Monday to announce their fall tour with original singer  \\
        20  &  The PG-13-rated thriller gave 20th Century Fox its first No. 1 launch in seven months. The opening-night crowd was heavily male and young, matching the video-game \\
        50 & Due to Wednesday night's victory, a mathematician and avid Vasco soccer fan calculated on Thursday that the team's chances of being dropped into the second division fell by\\
        90 & A top Iranian minister who admitted to faking his university degree will face a motion of no confidence on Tuesday on charges that he tried to bribe members of Parliament \\
        \bottomrule
      \end{tabular}
    \end{tiny}
  \end{center}
  \vskip -0.1in
\end{table*}

\begin{table}[t]
  \caption{Results on predictive document hashing. For each model we show the representation dimension, number of distinct codes (i.e., clusters) induced on 208808 training documents, and top-100 precision on the test set.}
  \label{tab:wdw}
  \vskip 0.15in
  \begin{center}
    \begin{small}
      \begin{tabular}{l|ccc}
        \toprule
        & Dim & \# Distinct Codes &  Precision \\
        \midrule
        BOW & 20000 & 208808 &  26.66 \\
        BMSH & 128 & 208004 &   75.77 \\
        DVQ & 128 &  208655 &  76.80 \\
        \midrule
        AMMI & 128 & \textbf{153123} & \textbf{79.14} \\
        \bottomrule
      \end{tabular}
    \end{small}
  \end{center}
    \vspace{-5mm}
\end{table}

\subsection{Predictive Document Hashing}

Unsupervised document hashing only considers a single variable $Y$ and does not test the conditional formulation $Y|X$.
Hence we introduce a new task, predictive document hashing, in which $(X, Y)$ represent distinct articles that discuss the same event.
It is clear that $I(X, Y)$ is large: the large uncertainty of a random article is dramatically reduced given a related article.

We construct such article pairs from the Who-Did-What dataset \citep{onishi2016did}.
We remove all overlapping articles so that each article appears only once in the entire training/validation/test data
containing 104404/8928/7326 document pairs.
We give more details of the dataset in the supplementary material.
Similarly as before, we use $20000$-dimensional TFIDF representations as raw input
and consider the task of compressing them into $m=128$ binary bits.
The quality of the binary encodings is measured by top-100 matching precision:
given a test article $y$, we check if the correct corresponding article $x$ is included in 100 nearest neighbors of $y$
under the encoding based on the Hamming distance.

\subsubsection{Models}

AMMI now consists of a pair of encoders $p^\psi_{Z|Y}$ and $p^\phi_{Z|X}$ as well as a variational prior $q^\theta_Z$,
which are respectively parameterized by order-$o$, order-$h$, and order-$r$ Markov models over $\myset{0,1}^m$.
Based on our findings in the previous section we use $o=h=0$ and $r=3$.
We train the model with Algorithm~\ref{alg:adv}.

To compare with VAEs, we consider a conditional variant which optimizes the ELBO objective under the joint distribution
$p_{XYZ}(x, y, z) = p_Y(y) p_{Z|Y}(z|y) p_{X|Z}(x|z)$ with an approximation of the true posterior $q_{Z|Y}(z|y) \approx p_{Z|XY}(z|x, y)$
which can be used as a document encoder.
In this setting, training for BMSH remains unchanged except that the the model predicts $X$ instead of $Y$ for reconstruction and uses conditional prior $p_{Z|Y}$ in the KL regularization term.
The DVQ model likewise simply predicts $X$ instead of $Y$ but loses its ELBO interpretation.
We tune hyperparameters of all models similarly as before.

\subsubsection{Results and Qualitative Analysis}

Table~\ref{tab:main} shows top-100 precisions on the test portion.
We see that AMMI again achieves the best performance in comparison to BMSH and DVQ.
We also report the number of distinct values of $z$ induced on the 208808 training articles (union of $X$ and $Y$).
We see that AMMI learns the most compact clustering which nonetheless generalizes best.

We conduct qualitative analysis of the document encodings by examining articles with increasing Hamming distance (i.e., semantic drift).
Table~\ref{tab:drift} shows illustrative examples.
The article about the O.J.~Simpson trial drifts to the Phil Spector trial, the Ted Stevens trial, and eventually other unrelated subjects in politics and economy.
The article about NASCAR drifts to other racing reports, cycling, movie stars and politics.

\section{Conclusions}

We have presented AMMI, an approach to learning discrete structured representations by adversarially maximizing mutual information.
It obviates the intractability of entropy estimation by making mild structural assumptions that apply to a wide class of models
and optimizing the difference of cross-entropy upper bounds.
We have derived a concrete instance of the approach based on Markov models and identified important practical issues
such as the expressiveness of the variational prior.
We have demonstrated its utility on unsupervised document hashing by
outperforming current best results.
We have also proposed the predictive document hashing task and showed that AMMI yields high-quality semantic representations.
Future work includes extending AMMI to other structured models and extending it to cases in which even cross entropy calculation is intractable.

\section*{Acknowledgements}

We thank David McAllester for many insightful discussions.

\bibliography{mmi_features}
\bibliographystyle{icml2020}

\appendix

\section{Forward Algorithm}

The forward algorithm is shown in Algorithm~\ref{alg:forward}.
We write $[m]$ to denote the set of $m$ integers $\myset{1,\ldots,m}$
and $\doubbr{A} = 1$ if $A$ is true and $0$ otherwise.

\begin{algorithm}[t]
  \caption{Forward}
  \label{alg:forward}
  \begin{algorithmic}
    \STATE {\bfseries Input:} $p(z_i|i, z_{i-o:i-1})$ for $z_{i-o:i} \in \myset{0,1}^{o+1}$ and $i \in [m]$
    \STATE {\bfseries Output:} For $z_{i-o:i-1} \in \myset{0,1}^o$ and $i \leq m$
    \begin{align*}
      \pi(z_{i-o:i-1}|i) = \sum_{\substack{\bar{z} \in \myset{0,1}^{i-1}:\; \bar{z}_{-j} = z_{i-j}\;\forall j \in [o]}} p(\bar{z}) 
    \end{align*}
    \STATE {\bfseries Runtime:} $O(m 2^o)$
    \\\hrulefill
    \STATE {\bfseries Base:} $\pi(z_{-o:-1}|i) \gets \doubbr{z_{-o:-1} = 0^o}$ for $i \leq 1$
    \STATE {\bfseries Main:} For $i=2 \ldots m$, for $z_{i-o:i-1} \in \myset{0,1}^o$,
    \begin{align*}
      \pi(z_{i-o:i-1}|i) &\gets p(z_{i-1}|i-1, z^{(0)}) \times \pi(z^{(0)}|i-1) \\
      &\hspace{1mm} + p(z_{i-1}|i-1, z^{(1)}) \times \pi(z^{(1)}|i-1)
    \end{align*}
    where $z^{(b)} = (b, z_{i-o}, \ldots, z_{i-2}) \in \myset{0,1}^o$
  \end{algorithmic}
\end{algorithm}

\section{Dataset Construction for Predictive Document Hashing}

We take pairs from the Who-Did-What dataset \citep{onishi2016did}. 
The pairs in this dataset were constructed by drawing articles from the LDC Gigaword
newswire corpus. A first article is drawn at random and then a list of candidate second articles is drawn using the
first sentence of the first article as an information retrieval query. 
A second article is selected from the candidates using criteria described in \citet{onishi2016did}, 
the most significant of which is that the second article must have occurred 
within a two week time interval of the first. 
We filtered article pairs so that each article is distinct in all data. 
The resulting dataset has 104404, 8928, and 7326 article pairs for training, validation, and evaluation.

To follow the standard setting in unsupervised document hashing, 
we represent each article as a TFIDF vector using a vocabulary of size 20000.
We use SentencePiece \citep{kudo2018sentencepiece} to learn an optimal text tokenization 
for the target vocabulary size.

\end{document}